%% file: main.tex
\documentclass[final]{cvpr}

\usepackage{times}
\usepackage{epsfig}
\usepackage{graphicx}
\usepackage{amsmath}
\usepackage{amssymb}
\usepackage{float}
\usepackage{mwe}
\usepackage{multirow}
\usepackage{diagbox}
\usepackage{enumitem}
\usepackage{sidecap}
\DeclareMathOperator*{\argmin}{arg\,min}
\usepackage{soul}

\usepackage[pagebackref=true,breaklinks=true,colorlinks,bookmarks=false]{hyperref}

\def\E{\mathbf{E}}
\def\D{\mathbf{D}}
\def\De{\mathbf{D_e}}
\def\Di{\mathbf{D_i}}
\def\I{\mathbf{I}}

\def\etc{etc\onedot}
\def\ie{i.e\onedot}
\def\eg{e.g\onedot}
\def\cf{cf\onedot}

\def\wrt{w.r.t.\xspace}

\begin{document}

\title{Beyond Image to Depth: Improving Depth Prediction using Echoes}

\author{Kranti Kumar Parida\textsuperscript{1} \hspace{2em} Siddharth Srivastava\textsuperscript{2} \hspace{2em} Gaurav Sharma\textsuperscript{1,3} \vspace{0.5em} \\
\textsuperscript{1 }IIT Kanpur \hspace{1.5em}
\textsuperscript{2 }CDAC Noida \hspace{1.5em}
\textsuperscript{3 }TensorTour Inc. \vspace{0.2em} \\
{\small \texttt{\{kranti, grv\}@cse.iitk.ac.in, siddharthsrivastava@cdac.in}}}
\maketitle


\input{abstract}
\input{introduction}
\input{related_work}
\input{approach}
\input{experiments}

\input{conclusion}
{\small
\bibliographystyle{ieee_fullname}
\bibliography{egbib}
}
\clearpage

\appendix

\input{supplementary/dataset_details}
\input{supplementary/network_architecture}

\input{supplementary/implementation_details}
\input{supplementary/evaluation_metrics}
\input{supplementary/qual_res}
\end{document}

%% file: abstract.tex
\begin{abstract}
We address the problem of estimating depth with multi modal audio visual data. Inspired by the ability of animals, such as bats and dolphins, to infer distance of objects with echolocation, some recent methods have utilized echoes for depth estimation. We propose an end-to-end deep learning based pipeline utilizing RGB images, binaural echoes and estimated material properties of various objects within a scene. We argue that the relation between image, echoes and depth, for different scene elements, is greatly influenced by the properties of those elements, and a method designed to leverage this information can lead to significantly improved depth estimation from audio visual inputs. We propose a novel multi modal fusion technique, which incorporates the material properties explicitly while combining audio (echoes) and visual modalities to predict the scene depth. We show empirically, with  experiments on Replica dataset, that the proposed method obtains $28\%$ improvement in RMSE compared to the state-of-the-art audio-visual depth prediction method. To demonstrate the effectiveness of our method on larger dataset, we report competitive performance on Matterport3D, proposing to use it as a multimodal depth prediction benchmark with echoes for the first time. We also analyse the proposed method with exhaustive ablation experiments and qualitative results. The code and models are available at \url{https://krantiparida.github.io/projects/bimgdepth.html}
\end{abstract}

%% file: introduction.tex
\vspace{-4ex}
\section{Introduction}
\label{sec:intro}

\begin{figure}
\centering
\includegraphics[width=\columnwidth]{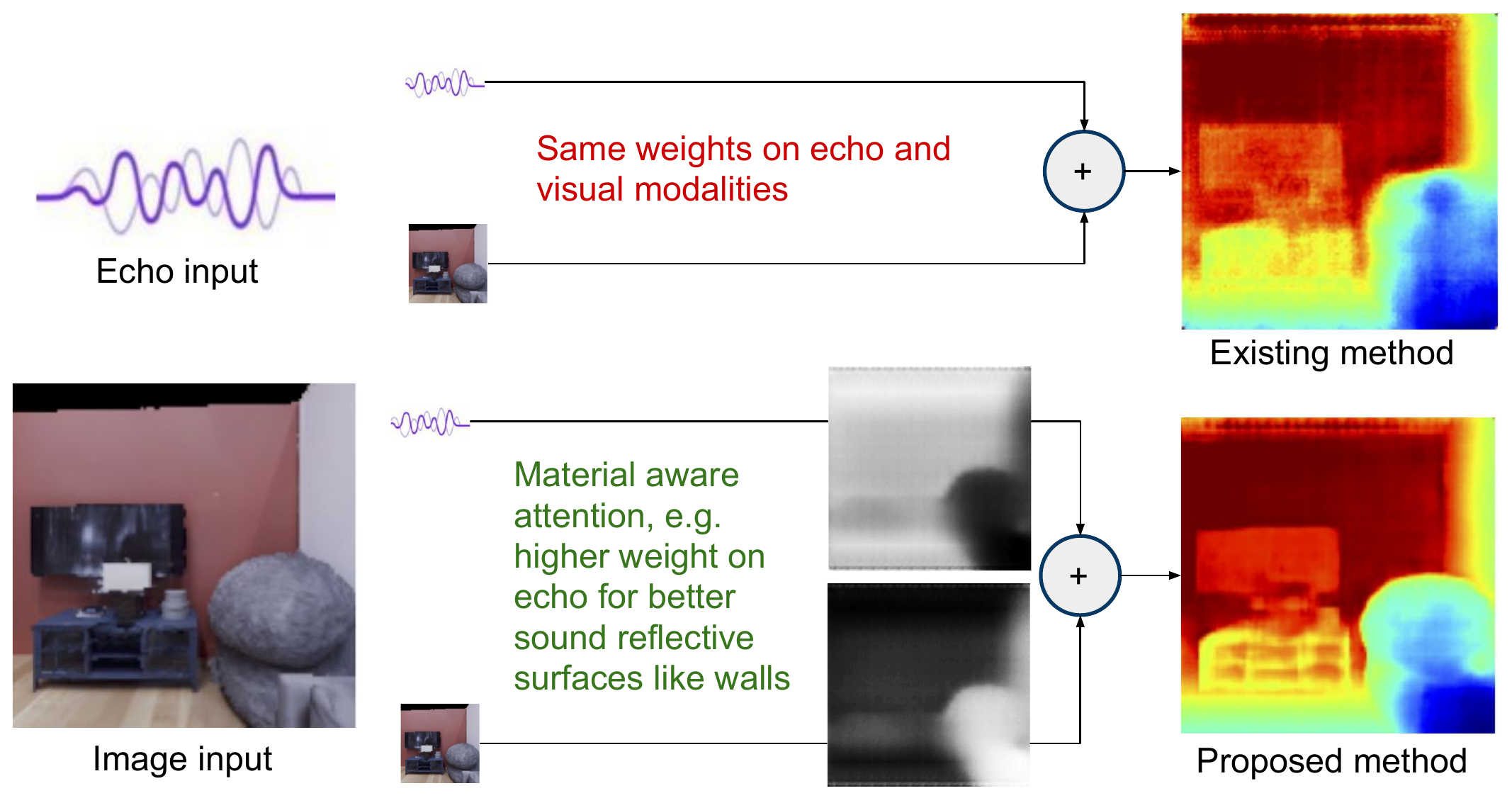}
\caption{
We address the problem of depth prediction using multimodal audio (binaural echo) and visual (monocular RGB) inputs. We propose an attention based fusion mechanisms, where the attention maps are influenced by automatically estimated material properties of the scene objects. We argue that capturing the material properties while fusing echo with images is beneficial as the light and sound reflection characteristics depend not only on the depth, but also on the material of the scene elements.
}
\label{fig:teaser}
\end{figure}

Humans perceive the surroundings using multiple sensory inputs such as sound, sight, smell and touch, with different tasks involving different combinations of such inputs.

In computer vision, multimodal learning has also gained interest. As one popular stream, researchers have leveraged audio and visual inputs for addressing challenging problems. These problems can be broadly divided into three categories: (i) using audio modality only as the input, to learn a seemingly visual task, e.g.\ using echo for depth prediction~\cite{christensen2019batvision}, (ii) using visual modality as \emph{auxilliary information} for an audio task, e.g.\ using videos to convert mono audio to binaural audio~\cite{gao20192}, and (iii) using both audio visual modalities together, \eg for depth prediction~\cite{gao2020visualechoes}. Here, we follow the third line of work, and address the problem of depth map prediction using both audio and visual inputs. Studies in psychology and perception indicate that both sound and vision complement each other, \ie visual information helps calibrate the auditory information ~\cite{kolarik2016auditory} while auditory grouping helps solve visual ambiguity~\cite{watanabe2001sound}. Many animals, like bats and dolphins, use echolocation to estimate the distances of objects from them. Visually impaired humans have also been reported to use echolocation \cite{HumanELWiki}. Motivated by such cases, Christensen et.~al \cite{christensen2019batvision, christensen2020batvision} recently showed that depth maps can be predicted directly from stereo sound. Gao et.~al \cite{gao2020visualechoes} showed that by fusing features from binaural echoes with the monocular image features, depth estimation can be improved. Inspired by these findings, we work with similar reasoning, \ie sound contains useful information to predict depth, and that echoes, used along with monocular images, improve depth estimation. 

Going beyond the current methods which do simple combinations of features from echoes and images \cite{gao2020visualechoes}, we argue that the material properties of the objects in the scene significantly inform the spatial fidelity of the two streams. Some objects may lend better depth estimates with echoes, while some may prefer the visual modality more. Deriving from this motivation, we propose a novel end-to-end learnable network with a multimodal fusion module. This novel module incorporates material properties of the scene and fuses the two modalities with spatial attention maps indicating the fidelity of the respective modality for different spatial locations. The material properties are automatically estimated using a sub-network initialized with training on auxiliary data on materials. As the final depth prediction, the method fuses the depth maps produced by the audio and visual inputs, modulated by the predicted attention maps. \figref{fig:teaser} illustrates the difference with a real output of an existing method and the proposed approach, showing qualitative improvements.

\begin{figure}
    \centering
    \includegraphics[width=\columnwidth]{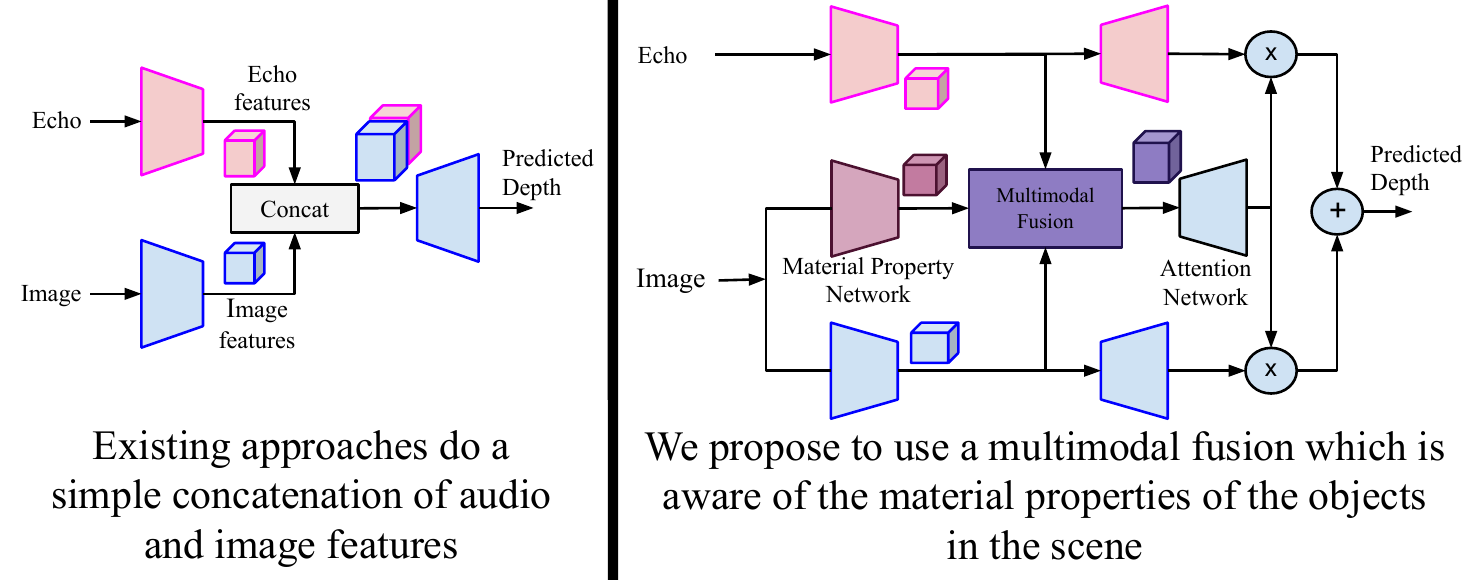}
    \caption{Comparison of our method with the existing approaches}
    \label{fig:comparison}
    \vspace{-1em}
\end{figure}

We demonstrate the advantages of the proposed method with experiments on Replica \cite{straub2019replica} and Matterport3D \cite{Matterport3D} datasets. We outperform the previous state-of-the-art on Replica dataset by $\sim28\%$ RMSE. On Matterport3D, which is more complex and larger ($5$x) than Replica, we provide results on the multimodal depth prediction task for the first time, and compare the proposed method with existing approaches and challenging baselines. We also show that the proposed network can estimate better depth with low resolution images. This is important in practical systems working on depth estimation from monocular images, as sensors capturing echoes can be used along with cameras, to not only enhance the performance of existing setup but also suffer lesser degradation in depth prediction with the reduction in the quality of images. Further, we give ablation experiments to systematically evaluate the different aspects of the proposed method.

In summary, we make the following contributions:
\begin{itemize}[leftmargin=1.25em, itemsep=-0.25em]
    \item We propose a novel end-to-end learnable deep neural network to estimate depth from binaural audio and monocular images.
    \item We provide exhaustive quantitative and qualitative results on Replica and Matterport3D datasets. On Replica, we outperform the previous state-of-the-art by $\sim 28\%$. On Matterport3D, we provide results benchmarking existing methods. The proposed method achieves state-of-the-art performance, outperforming the existing best method on Matterport3D by $\sim 4\%$.
    \item We provide exhaustive ablation experiments on the design choices in the network, and validate our intuitions with representative qualitative results.
\end{itemize}

%% file: related_work.tex
\section{Related Works}
\label{sec:related_work}
\noindent\textbf{Audio-visual learning.}
Recently there has been a surge in interest in audio-visual learning. In one line of work, the correspondence between both the modalities are used to learn the representation in each of the individual modality, in a self-supervised manner \cite{arandjelovic2017look, Arandjelovic_2018_ECCV, hu2020discriminative, morgado2020learning, owens2018audio}. In \cite{arandjelovic2017look, Arandjelovic_2018_ECCV}, the authors used an auxiliary task, of predicting whether the audio and image pair correspond to each other, to learn representations in each of the modalities. In \cite{owens2018audio}, the authors predicted if the audio and the video clip are temporally synchronized to learn representations. In \cite{hu2020discriminative}, the authors advanced a step further and localized sound generating objects in the image by leveraging the correspondence between audio and the image. In one of the recent approach, the authors in \cite{morgado2020learning}, have used the spatial correspondence between $360^{\circ}$ video and audio. In another line of work, the integration of both audio and video modality was done to increase the performance. Recently a variety of task such audio source separation \cite{zhao2018sound, zhao2019sound, gan2020music, gao2018learning, gao2019co}, zero-shot learning \cite{parida2020coordinated, mazumder2020avgzslnet}, saliency prediction \cite{tsiami2020stavis}, audio spatialization \cite{gao20192, morgado2018self} have used the information from both audio and video modalities to improve the performance cf.\ using single modality only.

\noindent\textbf{Depth estimation without echoes.} The depth estimation methods span from only monocular image based methods to multi modal methods. Usually, the modalities are sparse depth maps, LiDAR point clouds, bird's eye views, and normal maps. Monocular depth estimation methods involve utilizing a single RGB image to estimate dense depth~\cite{zhao2020monocular, bhoi2019monocular, tiwari2020pseudo}. Many methods directly utilize single image~\cite{monodepth2, li2019learning, Ranftl2020} or estimate an intermediate 3D representations such as point clouds~\cite{weng2019monocular,you2019pseudo} and bird's eye views~\cite{srivastava2019learning} to estimate dense depth maps. A few other methods work on combining RGB with sparse depth maps, normal maps \etc~\cite{qiu2019deeplidar, ma2019self} to estimate dense depth maps. 

\noindent\textbf{Depth estimation with echoes.}
In \cite{christensen2019batvision, christensen2020batvision} depth of the scene was estimated using only echoes received from a single audio pulse. This approach completely ignored the visual modality while estimating the depth. On similar lines, the authors in \cite{vasudevan2020semantic} estimated the depth of scene directly from binaural audio of the object itself. They did not have the ground truth depth map, and instead used a vision network to predict the ground truth depth map. This method although used the direct audio from the object but the performance of the system was always upper bounded by the predicted depth map from visual input. In all of the methods above, the authors used one modality in isolation and did not fuse multi-modal information to improve the performance of the system. In \cite{gao2020visualechoes} the authors 
used echolocation as a pre-training task for learning a better visual representation. The authors 
also gave a case study, where they 
showed that simply concatenating the audio features with the visual features improves depth prediction. We explore the idea further, of improving depth prediction by adding binaural echoes to image input as well, and propose a novel multimodal fusion method which incorporates automatically estimated material properties in addition. We give the comparison with existing approach in \figref{fig:comparison}.

%% file: approach.tex
\begin{figure*}
    \centering
    \includegraphics[width=\textwidth]{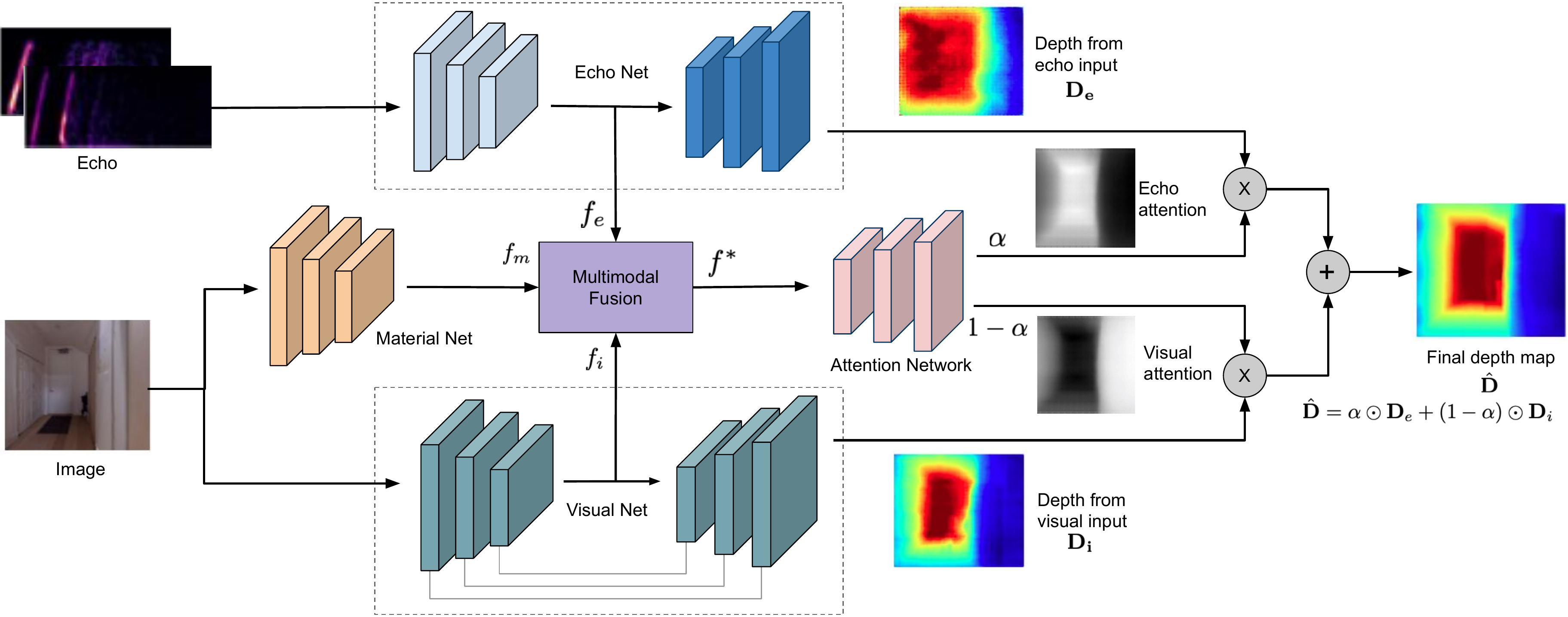}
    \caption{\textbf{Block diagram showing the overall architecture of the proposed method.} The network predicts depth independently from binaural echoes and image using Echo Net and Visual Net respectively. The Material Net gives the material features and the multimodal fusion block takes the image feature ($f_i$), material feature ($f_m$) and echo feature ($f_e$) as input and fuses them into a combined feature ($f^*$). The combined feature ($f^*$) is then input to attention network to get point-wise attention maps for the echo and image based depths, respectively. The final depth is predicted by taking a weighted combination of echo and image depth with attention maps as the weights.}
    \label{fig:block_diagram}
    \vspace{-1em}
\end{figure*}

\section{Approach}
\label{sec:approach}
We are interested in using echoes for predicting depth map of a scene. There is a spatio-temporal relationship between the received echoes and the depth of the scene, \ie the echoes received at different instant of time directly relate to the different depths in the scene. Let $x(t)$ be the original audio signal and $y(t)$ be the echo response of the scene. Assuming we have $d_i$  distinct depths of materials $m_i$ in the scene (discretized and grouped for simplicity of discussion), the obtained echo response can be approximated with a summation of time delayed original signal, reflecting off the materials at different depths. The amplitudes of the delayed signals will depend upon the material they hit respectively. Considering only first order echoes, say every distinct $d_i$ object contributes to a time delay of $t_i$, and the corresponding material changes the amplitude of the signal by a factor of $a_i$ on an average. The final response\footnote{The final response will include the original signal as well, as the sound emitter and recorder are both turned on together for a brief period of time} could then be approximated as
\begin{align}
    y(t)
     &= x(t) + \sum_{i=1}^{k} a_i x(t-t_i).
     \label{eq:reconstruction}
\end{align}

With $v_s$ denoting the speed of sound in the medium, the time delay can be directly associated with depths as $t_i = \frac{2 d_i}{v_s}$. Further, the amplitudes, $a_i$ of each time-delayed signal, are dependent on the acoustic absorption and reflection coefficient of the material.

Hence, the goal of making the network learn depth from the received echo, is influenced by two factors, (i) the 
relationship between the echoes and spatial depth variation in the scene, and (ii) the different acoustic properties of the scene objects. We propose a carefully tailored attention mechanism (\secref{sec:attn}) between the image and audio modalities for addressing the spatial variation aspect. In addition we also propose to incorporate material property estimation (\secref{sec:matnet}) as a proxy, to account for different properties of the scene elements which inform sound and light reflection and absorption, and hence inform the final depth prediction.

\subsection{Overall architecture}
We show the block diagram of the proposed network in \figref{fig:block_diagram}. The network consists of the following components: (i) echo subnetwork, (ii) visual subnetwork, (iii) material properties subnetwork, (iv) the multimodal fusion module and finally (v) attention prediction subnetwork. The echo and visual subnetworks consist of encoder-decoder pairs which estimate depth maps of the scenes independently. We input three feature maps coming from echo, visual and material property subnetworks respectively to the multimodal fusion network. The multimodal fusion module produces the fused features which we then feed to the attention network. We further use the attention network to predict two attention maps, one each for the two depth maps obtained from echo and visual decoder networks respectively. We then combine the individual depth maps using these attention maps to output the final depth map of the scene. We now give the details of the different components below. 
Please also see the supplementary document for detailed layer-wise architecture of the method.

\subsection{Echo Net for Echo to Depth} \label{subsec:echonet}
The echo net is an encoder decoder network which predicts depth from binaural echo input. We convert the time-domain echo response into a frequency domain spectrogram representation, $\E \in \mathbb{R}^{2\times P \times Q}$, where $P$ is the number of discrete time steps and $Q$ is the number of frequency bins. We input the spectrogram to the encoder part of the network to obtain the encoded representation $f_e \in \mathbb{R}^{N}$ of the echo, which is also one of the inputs to the multimodal fusion module. We then reshape the encoded vector to $N \times 1 \times 1$ and feed it to a series of fractionally strided convolution layers to get the depth map, $\De \in \mathbb{R}^{W \times H}$, where $W$, $H$ are the width and the height of the input image. While the upsampling happens here from an extreme $1\times1$ feature, \ie there is practically no spatial information (except what might get coded in the feature vector itself), such depth prediction from audio has been reported by earlier works also with fair success \cite{gao2020visualechoes}.

\vspace{-0.25em}
\subsection{Visual Net for Image to Depth}\label{subsec:visualnet}
The visual net is also an encoder decoder network which predicts depth from monocular RGB image. The architecture of this network is inspired from U-Net \cite{ronneberger2015u}, and consists of regular convolutional and fractionally 
strided convolutional layers with skip connections between them. We give the image, 
$\I \in \mathbb{R}^{3 \times W \times H}$ as input to the network which predicts the depth map, $\Di \in \mathbb{R}^{W \times H}$. We also use it to obtain the visual features from the intermediate layer (output of last conv layer) of the network denoted, 
$f_i \in \mathbb{R}^{N \times w \times h}$. We use this feature as one of the inputs to the multimodal fusion module as well.

\vspace{-0.5em}
\subsection{Material Net for Material Properties}\label{subsec:matnet}
\label{sec:matnet}
We use this network to extract the material properties of the objects present in the scene. We use a ResNet-18 architecture \cite{he2016deep} and feed the RGB image, $\I \in \mathbb{R}^{3 \times W \times H}$ as the input. We obtain a feature map, $f_m \in \mathbb{R}^{N \times w \times h}$ which encodes the material properties over the spatial locations in the scene image. This feature is the third input to the multimodal fusion module. We initialize the material network with pretraining on a large materials dataset \cite{bell2015material} with classes such as fabric, brick, asphalt, wood, metal, and then train it end to end with the rest of the network. We expect this initial material encoding capability of the network to be a proxy for encoding properties related to sound and light absorption and reflection, which affect depth prediction. Although the network evolves with the end to end training, the attention maps obtained qualitatively validate our assumptions (\secref{subsec:qual}).

\vspace{-0.5em}
\subsection{Multimodal Fusion Module}
The multimodal fusion module combines features from the three sources discussed above, \ie echo $f_e \in \mathbb{R}^{N}$, visual $f_i \in \mathbb{R}^{N \times w \times h}$ and material $f_m \in \mathbb{R}^{N \times w \times h}$. Given the motivation, discussed in \secref{sec:intro}, that different object might give different depth prediction performances with audio or visual modalities respectively, the multimodal fusion module helps us combine the modalities to provide as input to the attention prediction network (\secref{sec:attn}). 

We perform two bilinear transforms on the features to obtain two fusion maps, $f_{img}^j$ and $f_{mat}^j$, where $j=1,2,...K$ is the number of output channels in the bilinear transformation,
\begin{align}
    f_{img}^j(p,q) &= f_e^T\mathbf{A}_{img}^jf_i(p,q)+b_{img}^j, \forall p,q \\
    f_{mat}^j(p,q) &= f_e^T\mathbf{A}_{mat}^jf_m(p,q)+b_{mat}^j, \forall p,q
\end{align}
where, $(p,q)$ indexes the spatial coordinates, $\mathbf{A}_{img}^j$, $\mathbf{A}_{mat}^j$ are learnable weights of dimension $N \times N$ and $b_{img}^j$, $b_{mat}^j$ are scalar biases.

We finally concatenate the fusion maps 
$f_{img} \in \mathbb{R}^{N \times w \times h}$ and $f_{mat}\in \mathbb{R}^{N \times w \times h}$ along the first dimension to get the final fusion map $f^*=concat(f_{img}, f_{mat})$ to be fed into the per-pixel attention network.

\subsection{Attention Network}
\label{sec:attn}
The attention network is the final part of the network which we use to predict the per-pixel attention map given the concatenated fusion maps obtained in the previous step. The network consists of a series of fractionally strided convolutional layers with a final \texttt{Sigmoid} layer to normalize the values in the range $[0,1]$. 

The output of the network is an attention map $\alpha \in \mathbb{R}^{1 \times W \times H}$. We use the attention map $\alpha$ for weighting the echo predicted depth map $\D_e$ and $1-\alpha$ for the image predicted depth map $\D_i$. The final depth map $\hat{\D}$ is thus,

\begin{equation}
    \hat{\D} = \alpha \odot \D_e + (1-\alpha) \odot \D_i
\end{equation}

where, $\odot$ denotes pointwise multiplication.

\subsection{Loss Function and Training}
We train the network following \cite{hu2019revisiting}, and
use the logarithm of depth errors. The loss is given as,
\begin{equation}
    \mathcal{L}(\hat{\D}, \D) = \frac{1}{W H} \sum_{p=1}^{W}\sum_{q=1}^{H}\ln (1+\lVert \D(p,q) - \hat{\D}(p,q)\rVert_1),
\end{equation}
where $D$ is the ground truth depth map.

The full optimization problem is given by
\begin{equation}
    \theta_e^*, \theta_i^*, \theta_a^*, \theta_f^*, \theta_m^* = \argmin_{\theta_e, \theta_i, \theta_a, \theta_f, \theta_m} \mathcal{L}(\hat{\D}, \D).
    \label{eq:final_loss}
\end{equation}

where, $\theta_e$, $\theta_i$, $\theta_a$, $\theta_f$, $\theta_m$ are the parameters for echo to depth network, image to depth network, attention network, fusion module and material property network respectively. We ignore the undefined regions in the ground truth depth maps, and therefore, such regions do not contribute to the learning. Adding smoothness constraints~\cite{li2019learning} can potentially further improve the quality of generated depth, however we obtain good results without using them here. We train the full network in an end to end fashion using standard backpropagation for neural networks. 

%% file: experiments.tex
\section{Experiments}
\label{sec:experiments}

\subsection{Implementation Details}
\noindent\textbf{Datasets.} We perform experiments on Replica \cite{straub2019replica} and Matterport3D~\cite{Matterport3D} datasets. Both the datasets contain indoor scenes. Replica has a total of $18$ scenes covering hotels, apartments, rooms and offices. Matterport3D contains $90$ scenes. 
On Replica, we follow \cite{gao2020visualechoes} and use $15$ scenes for training and $3$ for testing. On Matterport3D, we use $77$ scenes for evaluation, out of which $59$, $10$ and $8$ scenes are used as train, validation and test respectively.  We simulate echoes on these datasets by using the precomputed room impulse response (RIR) provided by \cite{chen2019soundspaces} using 3D simulator Habitat~\cite{savva2019habitat} which takes into consideration the scene geometry and the materials present in the scene. We obtain the echoes by convolving input audio signal with the RIR. We use the material recognition dataset MINC~\cite{bell2015material} for pre-training the material net. 

\noindent\textbf{Network Architecture.} 
We use the same architecture for echo encoder, and image to depth encoder and decoder (Visual Net) as \cite{gao2020visualechoes}, for fair comparison and demonstrating the effectiveness of the proposed material and attention networks. We use the first four convolutional layers of ResNet-18 for the material property network. We initialize them with pretraining on ImageNet and MINC dataset. 

\noindent\textbf{Input Representation.} The input to Visual Net and Material Net is a $128 \times 128$ RGB image. For input to Echo Net, we use the spectrogram of $60$ms echo signal. For training on Replica, we use a sampling frequency of $44.1$ kHz and for Matterport3D, we use a sampling frequency of $16$ kHz. We use Hanning window of length $64$ and $32$ to compute spectrogram for Replica and Matterport3D respectively. We use FFT size of 512 for both the cases.

\noindent\textbf{Metrics.} Following earlier works in depth estimation, we report results on root mean squared error (RMSE), mean relative error (REL), mean $log_{10}$ error, and the percentage $\delta_t$ of pixels with both the relative error and its inverse under threshold $t$ where $t \in \{1.25, 1.25^2, 1.25^3\}$.

Due to space constraints, we provide more details on the datasets, network architecture, parameter settings and evaluation metrics in the supplementary material.

\subsection{Experiment Design}
We design the experiments below to demonstrate the following points. (i) Using audio and visual modalities together improves performance over either of them. (ii) Using material properties in addition improves further. (ii) Among the different ways to combine the three, \ie visual, audio and material properties, the proposed attention based fusion performs the best. We demonstrate the first two points with ablation experiments where we combine the inputs by simple concatenation, followed by a decoder to predict the depth map (\secref{sec:ablation} first part). Further, we demonstrate the third point by comparing combination methods and showing that attention based combination performs the best (\secref{sec:ablation} second part).

We then compare our full method with existing state of the art approaches (\secref{sec:soa}). We also show experiments on degrading resolution of image (\secref{sec:res}).

\subsection{Ablation Study}
\label{sec:ablation}

\noindent\textbf{Combination of echo, image and material properties.}
We show the results of combining the three inputs with simple concatenation in \tabref{tab:material_property_with_echo}. 
With only binaural echo as input, the RMSE is $0.995$, which improves to $0.673$ when image is added as well. 
When material property features are used with echo, the RMSE improves to $0.523$ \ie an improvement of $\sim47\%$ over echo only input and $\sim22\%$ over image+echo input. 

Lastly, when image and material property features are concatenated with echo features (all), the RMSE value further improves to $0.491$ \ie $\sim50\%$ over echo only input and $\sim27\%$ over echo+image input. 

These experiments validate that even with simple fusion, material properties improve the performance of the system. We attribute the improvement to our intuition that adding material properties explicitly allows the network to internally module the audio and visual features.

In the following we demonstrate the proposed explicit multimodal fusion followed by attention based weighting performs much better than the simple concatenation.

\begin{table}
    \centering
    \resizebox{\columnwidth}{!}{%
    \begin{tabular}{c|c|c|c|c|c|c}
    \hline
    Modality & RMSE ($\downarrow$) & REL ($\downarrow$) & log10 ($\downarrow$) & $\delta_{1.25}(\uparrow)$&$\delta_{1.25^2}(\uparrow)$&$\delta_{1.25^3}(\uparrow)$\\
    \hline \hline
    echo & 0.995 & 0.638& 0.208& 0.388& 0.599&0.742 \\
    \hline
    echo+img & 0.673& 0.451& 0.146 &0.534 &0.734 &0.845\\
    echo+mat. & 0.523& 0.282& 0.103& 0.652& 0.839& 0.920\\
    all& \textbf{0.491}& \textbf{0.276}& \textbf{0.098}& \textbf{0.667}& \textbf{0.846}& \textbf{0.924}\\
    \hline
    \end{tabular}
    }
    \caption{\textbf{Depth estimation by combining different modalities} Using echoes only (echo), echoes with image features (echo+img.), echoes with material features (echo+mat.) and combination of echo, image and material features (all). $\downarrow$ indicates lower is better and $\uparrow$ indicates higher is better. 
    }
    \label{tab:material_property_with_echo}
    \vspace{-0.5 em}
\end{table}

\noindent\textbf{Impact of multimodal fusion and attention.} 
\label{sec:exp_fusion}
We now validate the efficacy of our audio visual fusion method, which uses a multimodal fusion module to predict attention over the modalities to combine them. We compare the proposed fusion method, denoted \texttt{bilinear} with two alternatives, \ie a simple concatenation of features denoted \texttt{concat}, and a dot product based fusion denoted \texttt{dot}. All these methods fuse the features and use them to estimate attention weights. We also compare by the fusion method of VisualEchoes \cite{gao2020visualechoes}, which fuses features with concatenation and uses them with a decoder to predict depth map, \ie it has no attention based fusion.

We show the results in \tabref{tab:fusion_ablation}. We observe that \texttt{bilinear}, with an RMSE of $0.249$, performs best among the compared methods, highlighting that the proposed fusion is better than the simple concatenation or dot product based fusion. We also observe that \texttt{concat} performs better than VisualEchoes \ie $0.259$ \cf $0.346$ RMSE. This indicates that attention maps (which are present in \texttt{concat} but absent in VisualEchoes) are important for better performance. 
 
\begin{figure}
    \centering
    \includegraphics[width=\columnwidth]{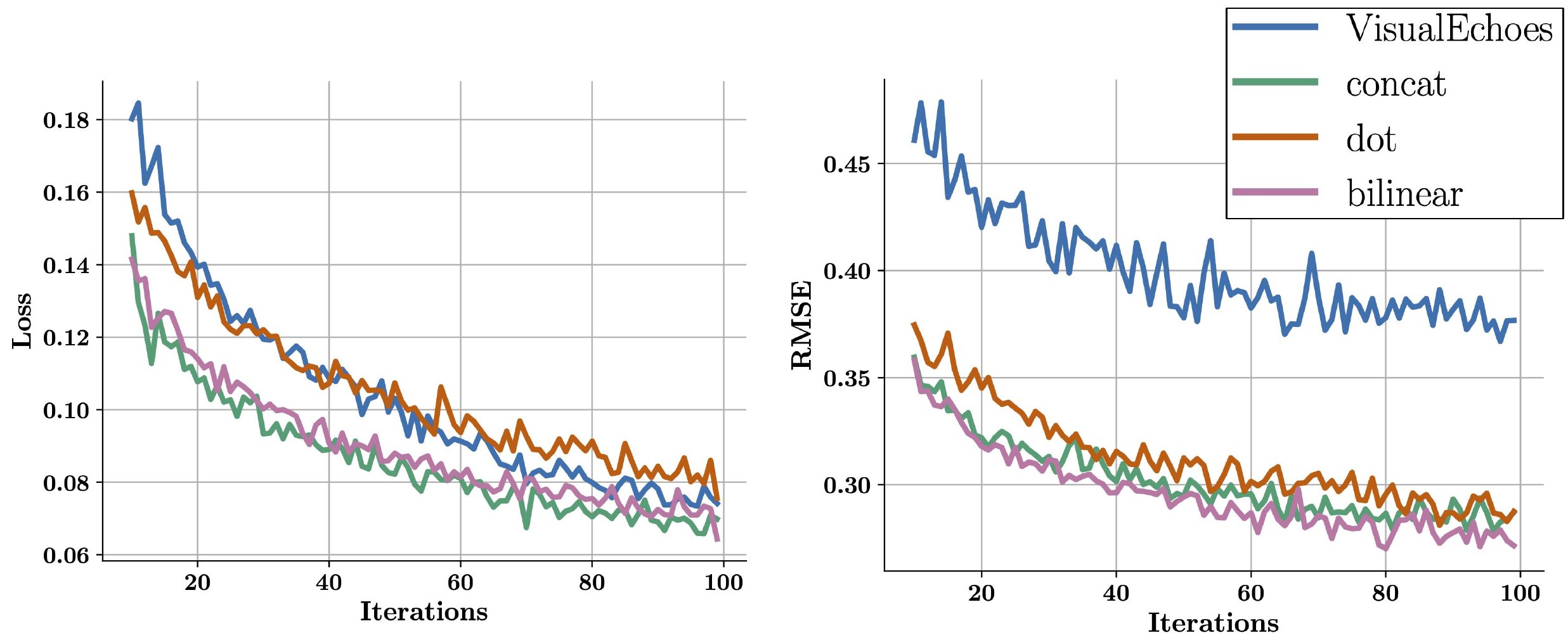}
    \caption{\textbf{Training loss and validation RMSE on Replica dataset.} In VisualEchoes \cite{gao2020visualechoes} depth prediction is performed directly from concatenated features. \texttt{dot}, \texttt{concat} and \texttt{bilinear} are the three different fusion strategies for proposed attention based prediction.}
    \label{fig:loss_rmse_plot}
    \vspace{-1em}
\end{figure}

\figref{fig:loss_rmse_plot} further shows the training loss (left) and validation RMSE (right) plots. We observe that VisualEchoes suffers from severe overfitting (much higher val RMSE), which is mitigated on adding the material features (\ie \texttt{concat}). This further reinforces the hypothesis that material properties play an important role in depth prediction. 

To conclude, we demonstrated from the ablation experiments that, (i) adding material properties explicitly is helpful for audio visual depth prediction, (ii) the proposed fusion strategy is better than simpler alternatives, and (iii) attention based combination of depth maps is better than simple concatenation as used in previous methods, \eg VisualEchoes.

\begin{table}
    \centering
    \resizebox{\columnwidth}{!}{%
    \begin{tabular}{c|c|c|c|c|c|c}
    \hline
    Method & RMSE ($\downarrow$) & REL ($\downarrow$) & log10 ($\downarrow$) & $\delta_{1.25}(\uparrow)$&$\delta_{1.25^2}(\uparrow)$&$\delta_{1.25^3}(\uparrow$)\\
    \hline \hline
    VisualEchoes \cite{gao2020visualechoes} & 0.346 & 0.172 & 0.068 & 0.798 & 0.905 & 0.950\\
    \hline
    \texttt{concat} &0.259 &0.122 &0.048 &0.867 &0.939 &0.968 \\
    \texttt{dot}& 0.262& 0.133& 0.050& 0.853& 0.943& \textbf{0.974}\\
    \texttt{bilinear}& \textbf{0.249}& \textbf{0.118}& \textbf{0.046}& \textbf{0.869}& \textbf{0.943}& \textbf{0.970}\\
    \hline
    \end{tabular}
    }
    \caption{\textbf{Performance of different fusion strategies}. \texttt{concat} refers to the concatenation of all the inputs, \texttt{dot} to fusion by dot product, and \texttt{bilinear} to fusion by bilinear transformation (see \secref{sec:exp_fusion}).}
    \label{tab:fusion_ablation}
    \vspace{-1.5 em}
\end{table}

\vspace{-0.5em}
\subsection{Comparison to state-of-the-art}
\label{sec:soa}
\noindent\textbf{Baselines.} We compare on Replica and Matterport3D against VisualEchoes~\cite{gao2020visualechoes} and competitive baselines. The baseline methods are AVERAGE, ECHO2DEPTH and RGB2DEPTH. AVERAGE refers to average depth value of all the samples in training set. ECHO2DEPTH refers to the depth estimation using only Echo Net (\secref{subsec:echonet}) and RGB2DEPTH refers to depth estimation using only Visual Net (\secref{subsec:visualnet}).

\noindent\textbf{Comparison on Replica dataset.} We report results in \tabref{tab:replica}. The proposed method outperforms all the compared methods on all the metrics. Specifically, it outperforms VisualEchoes by $\sim28\%$ on RMSE. We also observe that while the improvement of VisualEchoes \wrt RGB2DEPTH is marginal ($0.346$ \cf $0.374$ \ie $7.4\%$), the proposed method is able to achieve an improvement of $\sim33\%$ ($0.249$ \cf $0.374$ RMSE). Both the methods perform significantly better than AVERAGE and ECHO2DEPTH baselines. 

\noindent\textbf{Comparison on Matterport3D dataset.} We report results in \tabref{tab:mp3d}. We outperform echo only (ECHO2DEPTH), image only (RGB2DEPTH) and AVERAGE baselines on all the five metrics. Our method also outperforms the existing VisualEchoes method by $\sim4\%$ on RMSE and on all the metric after training the method for $300$ epochs. Further, better results on $\delta$ indicate that the proposed method has lower pixel wise relative error \cf VisualEchoes which are manifested in form of better depth estimation around edges (\secref{subsec:qual}). 

Since Matterport3D is a popular benchmark for depth estimation, we also compare our method with the state-of-the-art methods on sparse to dense depth map estimation. These methods use sparse depth maps as inputs, while we have no explicit depth information in the inputs. We also use a slightly smaller subset of Matterport3D, \ie $77$ \cf $90$ scenes for other. The results are shown in~\tabref{tab:mp3d_sparse_depth} where we obtain better results than four out of five compared methods. While the performances are not directly comparable, it supports the argument that echo can be a viable modality for estimating depth from RGB and can potentially provide additional information that are usually obtained from explicit 3D representations such as sparse depth maps.

\begin{table}
    \centering
    \resizebox{\columnwidth}{!}{%
    \begin{tabular}{c|c|c|c|c|c|c}
    \hline
    Method & RMSE ($\downarrow$) & REL ($\downarrow$) & log10 ($\downarrow$) & $\delta_{1.25}$ ($\uparrow$) & $\delta_{1.25^2}$ ($\uparrow$) & $\delta_{1.25^3}$ ($\uparrow$)\\
    \hline \hline
         AVERAGE & 1.070 & 0.791 & 0.230 & 0.235 & 0.509 & 0.750 \\
         ECHO2DEPTH & 0.713 & 0.347 & 0.134 & 0.580 & 0.772 & 0.868 \\
         RGB2DEPTH & 0.374 & 0.202 & 0.076 & 0.749 & 0.883 & 0.945\\
         VisualEchoes \cite{gao2020visualechoes} & 0.346 & 0.172 & 0.068 & 0.798 & 0.905 & 0.950\\
         Proposed Method& \textbf{0.249}& \textbf{0.118}& \textbf{0.046}& \textbf{0.869}& \textbf{0.943}& \textbf{0.970}\\
         \hline
    \end{tabular}
    }
    \caption{\textbf{Comparison with existing methods on Replica dataset}. We report the results for the baseline and existing methods directly from \cite{gao2020visualechoes}.}
    \label{tab:replica}
\end{table}

\begin{table}
    \centering
    \resizebox{\columnwidth}{!}{%
    \begin{tabular}{c|c|c|c|c|c|c}
    \hline
    Method & RMSE ($\downarrow$) & REL ($\downarrow$) & log10 ($\downarrow$) & $\delta_{1.25}$ ($\uparrow$) & $\delta_{1.25^2}$ ($\uparrow$) & $\delta_{1.25^3}$ ($\uparrow$)\\
    \hline \hline
         AVERAGE & 1.913& 0.714& 0.237& 0.264& 0.538& 0.697\\
         ECHO2DEPTH & 1.778& 0.507& 0.192& 0.464& 0.642& 0.759\\
         RGB2DEPTH & 1.090& 0.260& 0.111& 0.592& 0.802&0.910 \\
         VisualEchoes \cite{gao2020visualechoes} & 0.998& 0.193& 0.083& 0.711& 0.878& 0.945\\
         Proposed Method& \textbf{0.950}& \textbf{0.175}& \textbf{0.079}& \textbf{0.733}& \textbf{0.886}& \textbf{0.948}\\
         \hline
    \end{tabular}
    }
    \caption{\textbf{Comparison with existing methods on Matterport3D dataset.}}
    \label{tab:mp3d}
\end{table}

\begin{SCtable}
    \centering
    \resizebox{0.6\columnwidth}{!} {
    \begin{tabular}{c|c|c}
    \hline
    Method & RMS ($\downarrow$) & MAE ($\downarrow$) \\
    \hline \hline
        AD \cite{liu2013guided} &1.653 & 0.610\\
        MRF \cite{harrison2010image} &1.675 & 0.618\\
        Zhang et al. \cite{zhang2018deep} & 1.316 & 0.461 \\
        Huang et al. \cite{huang2019indoor} & 1.092 & 0.342 \\
        Xiong et al. \cite{xiong2020sparse} & 0.860 & 0.462\\
        \hline
        \emph{Proposed Method$^*$} & \emph{1.008} & \emph{0.570} \\
         \hline
    \end{tabular}
    }
    \caption{\textbf{Comparison on Matterport3D}. $^*$The compared methods use sparse depth maps as inputs, while we do not.}
    \label{tab:mp3d_sparse_depth}
\end{SCtable}

\vspace{-0.5em}
\subsection{Qualitative Results}\label{subsec:qual}
\begin{figure}
    \centering
    \includegraphics[width=\columnwidth]{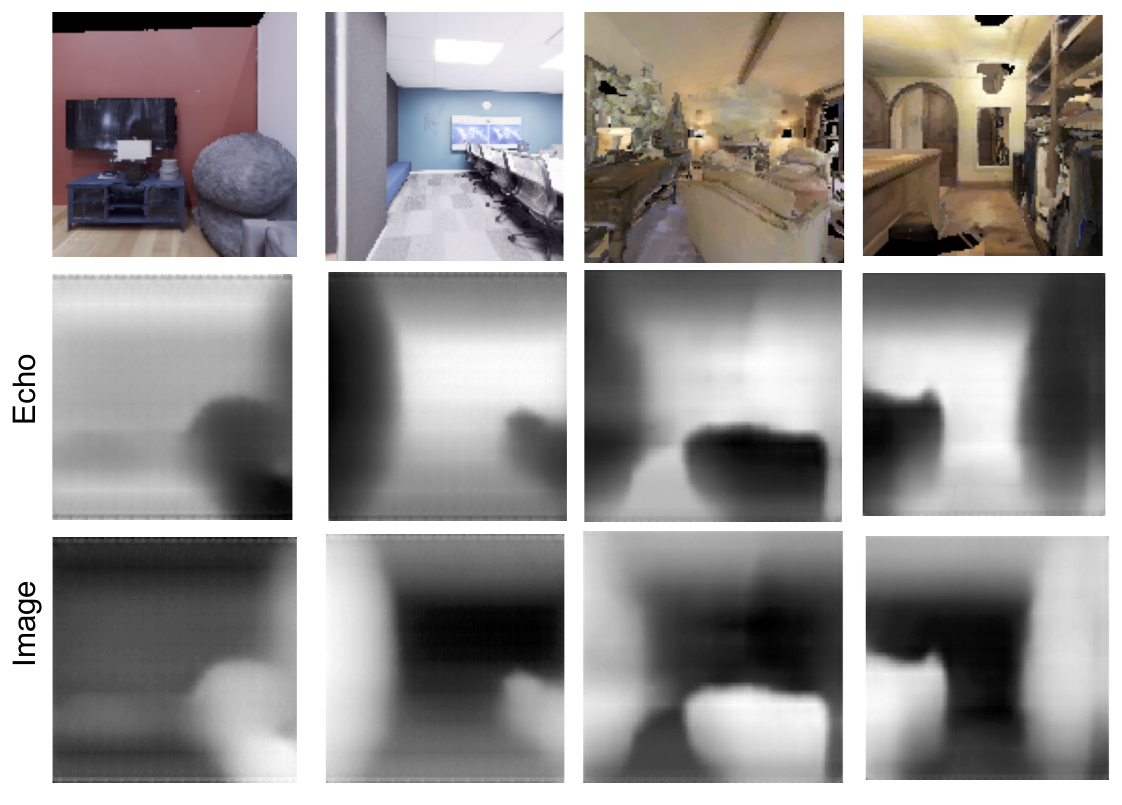}
    \caption{\textbf{Visualizing attention map.} We show attention maps for echo and image. First two columns are examples from Replica dataset and last two columns are examples from Matterport3D dataset. We observe that the echo modality, in general, produces high attention value for far away solid structure whereas the image modality attends more to nearby points (sofa in first and third example). See supplementary material for more qualitative results.}
    \label{fig:attention}
    \vspace{-1em}
\end{figure}

\begin{figure*}
    \centering
    \includegraphics[width=\textwidth]{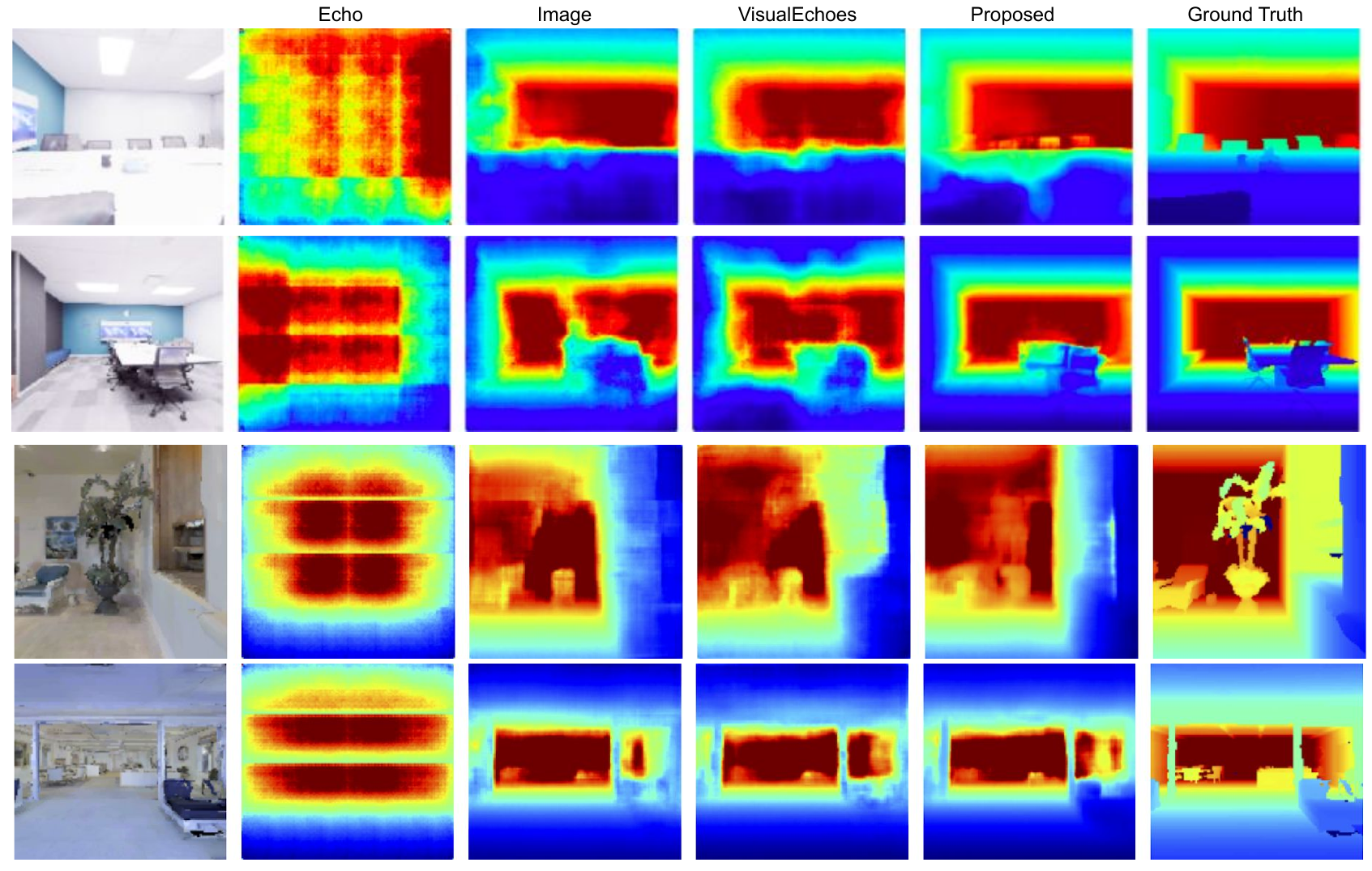}
    \caption{\textbf{Qualitative comparisons for depth estimation on Replica (first two rows) and Matterport3D (last two rows) datasets.} We observe that the proposed approach is able to preserve fine structures, and has better depth estimation at the boundaries when compared to the existing approach. See supplementary material for more qualitative results.}
    \label{fig:depth_pred}
    \vspace{-1.5em}
\end{figure*}

We give qualitative results of both (i) depth prediction, and (ii) attention maps in this section. We first provide few qualitative examples for attention map of echo and image respectively in Fig.~\ref{fig:attention}. We provide the attention maps for two examples each from Replica and Matterport3D. We make a general observation from these results that the echo attention map mostly looks into the far off regular regions but completely ignores the finer structures in them. It gives higher weight to the walls in the scenes but ignores any objects present on it (e.g.\ monitor on third example and wall painting on fourth). It also ignores the chairs and sofa which are present on relatively lower depth values. 

The image attention map, being complementary to the echo attention, gives higher weights to nearby objects such as the sofa, chair \etc, but ignores the far off wall in all the examples. This could be mainly due to two reasons (i) the echoes from the nearby regions are observed almost instantaneously which is hard to distinguish from the original signal (ii) most of the nearby objects (e.g.\ sofa) are made up of sound absorbing materials which do not reflect audio signals strongly. This results suggest that our network tries to leverage the best of both worlds, to get the final depth map.

\begin{table}
    \centering
    \resizebox{\columnwidth}{!}{%
    \begin{tabular}{c|c|c|c|c|c|c}
    \hline
    \diagbox{Method}{Scale} & $1$x & $\frac{1}{2}$x & $\frac{1}{4}$x & $\frac{1}{8}$x & $\frac{1}{16}$x & $\frac{1}{32}$x\\
    \hline \hline
        Img Only & 0.374	& 0.362	& 0.375	& 0.398	& 0.440	& 0.550 \\
        VisualEchoes\cite{gao2020visualechoes}&0.346&	0.354&	0.357&	0.392&	0.471&	0.593\\
        Proposed Method & 0.249	&0.244&	0.249&	0.281&	0.342&	0.446\\
        \hline
    \end{tabular}
    }
    \caption{\textbf{Performance on varying input image resolution.} The metric is RMSE (lower is better).}
    \label{tab:low_res_comparison}
    \vspace{-1.5em}
\end{table}

We give a few examples of reconstructed depth using our approach and also compare it with an existing approach in Fig.~\ref{fig:depth_pred}.  We observe that our approach is quite robust to fine grained structure in the scene \cf VisualEchoes \cite{gao2020visualechoes}. In the first example, VisualEchoes misses the small chairs in the scene while our method gives an idea of the structure. Again the boundaries in the first example are not very clearly identified by VisualEchoes, but are estimated almost perfectly in our case. We also observe similar trends in other three examples as well. The results from the individual modalities (image and echo) are not satisfactory, but do capture the overall structure of the scene. These results suggests that the networks are not individually powerful but their performance improves significantly when we effectively combine the information from both of them. We encourage the readers to look at the supplementary material for more such qualitative results.

\subsection{Experiments with varying image resolutions}\label{sec:res}
As discussed in \secref{sec:ablation}, the network can better exploit the combination of echo and image due to attention based integration. This motivates us to study the effect of resolution degradation. This is similar to the case where human vision degrades and the brain learns to compensate and adapt based on the auditory input~\cite{berry20143d, thaler2016echolocation}. We evaluate the proposed method by progressively degrading the image resolution. We give the results in ~\tabref{tab:low_res_comparison} by gradually reducing the resolution to $\frac{1}{2}$, $\frac{1}{4}$,$\frac{1}{8}$, $\frac{1}{16}$ and $\frac{1}{32}$ times the original image. 

We observe that the proposed approach is more robust to reduction in resolution of the images than the compared methods. The performance of image only method degrades significantly when the downscaling factor is $\frac{1}{8}$ of the original image size,  while the proposed method still performs better than image only method with original resolution \ie $0.281$ RMSE for proposed method at $\frac{1}{8}$x scale \cf $0.374$ with image only at $1$x.  Further, we observe that even with very high downscaling of $\frac{1}{32}$x, we obtain a better RMSE as compared to VisualEchoes\ ($0.446$ \cf $0.593$). In fact, VisualEchoes performs worse than even the image only method. Similar observations can be made at $\frac{1}{16}$x. We can also observe that the rate of decrease in RMSE from $1$x to $\frac{1}{4}$x is more in VisualEchoes \cf the image only and the proposed method. This further highlights the efficacy of the proposed method.

%% file: conclusion.tex
\vspace{-1em}
\section{Conclusion}
\vspace{-1ex}

\label{sec:conclusion}
We presented a novel method for estimating depth by combining audio and visual modalities. 
We hypothesised that material properties play a significant role in the task, and proposed to use automatic material property estimation to predict spatial attention maps which modulate and combine the outputs of audio and image based depth prediction. We showed with quantitative experiments on challenging benchmark datasets, that (i) adding material properties explicitly improves the depth prediction over audio and visual prediction, (ii) having an attention mechanism based fusion method is better than other simpler existing approaches for audio visual fusion. 
We also demonstrated qualitatively that the attention maps focus on interpretable areas for the two modalities. While audio attention maps tended to ignore materials which would diffuse or absorb the audio wave, the image based attention included those areas. We also demonstrated qualitatively that the proposed method performs better than existing method, especially near the depth edges and brings out the finer structures in the scene.

We further showed experiments with reduced image resolution where our method degraded gracefully, while the compared methods loses performances significantly. We even compared our method with existing methods for sparse to dense depth prediction, and obtained encouraging competitive results, while not using sparse dense data as input for our method. We would like to explore such multimodal fusion with other modalities like sparse point clouds in the future to obtain even higher quality depth predictions. Further, geometric prior~\cite{srivastava2021} can also be leveraged to improve the results.  

In conclusion, we believe that using echo for depth prediction, especially in combination with other modalities is a promising direction, especially given the low cost and wide availability of audio sensors.

\vspace{0.1em} 
\noindent
\textbf{Acknowledgment.}  Kranti Kumar Parida gratefully acknowledges support from the Visvesvaraya fellowship.

%% file: supplementary/dataset_details.tex
\section{Dataset Details}
\label{sec:dataset_details}

We use two datasets Replica \cite{straub2019replica} and Matterport3D \cite{Matterport3D} for our experiments. Both the datasets are rendered using an open source 3D simulator, Habitat \cite{savva2019habitat}. To obtain echoes on both the datasets, we use the simulations from Soundspaces~\cite{chen2019soundspaces}. Soundspaces augments the simulator by providing realistic audio simulations for the scenes by considering room geometry and materials in the room. 

\subsection{Simulating Echoes} We use the procedure outlined below to obtain echoes on both Replica and Matterport3D dataset. Soundspaces performs acoustic simulation in two steps as follows.

\noindent\textbf{Step 1.} The visual scene from the respective dataset is subdivided into grids. The grids are divided along navigable points so that an agent can be placed there. Then the Room Impulse Response (RIR) is computed between each pair of points using audio ray tracing \cite{veach1995bidirectional}. Each pair denotes a combination of source and receiver which send the audio signal and receive the echoes respectively. 
\\
\noindent\textbf{Step 2.}  The echoes are obtained by convolving the input audio signal with the RIR computed in the previous step. 

Following Soundspaces, we use the RIR between each pair of point at four orientations ($0^\circ$, $90^\circ$, $180^\circ$, $270^\circ$). For the proposed method, we place the source and receiver at the same point and use the resulting RIR.  In addition, following \cite{gao2020visualechoes}, we use the source audio signal as a $3$ ms sweep signal spanning the human hearing range ($20$Hz to $20$kHz). We obtain the echo response by convolving the corresponding source audio signal with the RIRs obtained previously. Further, the sampling rate for the source and received audio (echoes) are $44.1$ kHz and $16$ kHz for Replica and Matterport3D respectively. 

\subsection{Visual Scenes} We now provide details on the scenes used from each dataset along with the train and test details. 

\noindent\textbf{Replica dataset.} We use all the $18$ scenes from Replica having $6960$ points in total, from $1740$ images and $4$ orientations.  Following \cite{gao2020visualechoes}, we use a train set consisting of $5496$ points and $15$ scenes. The test set consists of $1464$ points from $3$ scenes. As a validation set is not defined for Replica, we use a small subset of points from train set for tuning the network parameters. Then the parameters are fixed, and the entire train set is used training the network. 

\noindent\textbf{Matterport3D dataset.} Matterport3D consists of $90$ scenes. Soundspaces provides RIR for $85$ of these scenes. Further, we discard another $8$ scenes which have none or a very few navigable points. This results in a dataset with $77$ scenes which we use as our final dataset. These $77$ scenes contain $67,376$ points from $16,844$ and $4$ orientations. The dataset is then split into train, validation and test sets. The train set consists of $40,176$ points and $59$ scenes. The validation set consists of $13,592$ points and $10$ scenes. The test set consists of $13,602$ points and $8$ scenes.  
 

%% file: supplementary/network_architecture.tex
\section{Network Architecture and Parameters}
\label{sec:network_architecture}

We now provide the detailed architecture of each subnetwork from the proposed method. 

\noindent\textbf{Echo Net.} It is an encoder-decoder network. The encoder is inspired from \cite{gao2020visualechoes} and consists of a convolutional neural network having $3$ layers with filter dimensions $8\times8$, $4\times4$, $3\times3$ and stride $4\times4$, $2\times2$ and $1\times1$ respectively for each layer. The number of output filters in each layers are $32$, $64$ and $8$ respectively. Finally, we use a $1 \times 1$ conv layer to convert arbitrary sized feature dimension into a $512$ dimensional feature vector. 

The decoder consists of $7$ fractionally strided convolutional layers with filter size, stride and padding of $4$,$2$ and $1$ respectively. The number of output filters for the $7$ layers are $512, 256, 128, 64, 32, 16$ and  $1$ respectively. We use BatchNorm and RELU non-linearity after each layer of the network.

\noindent\textbf{Visual Net.} It consists of an encoder decoder network. The encoder consists of a convolutional neural network with $5$ layers. For each layer, the filter size is $4$, the stride is $2$ and padding is $1$. The $5$ layers have $64, 128, 256, 512, 512$ number of output filters respectively. We use LeakyRELU with negative slope of $0.2$ and BatchNorm after each layer. 

Similarly for the decoder we use $5$ fractionally strided convolutional layers with output filters of size $512, 256, 128, 64, 1$ respectively. We also use skip connections and concat the features from the corresponding encoder layer with decoder output to get the final feature map from the decoder. We use BatchNorm and RELU after each layer.

\noindent\textbf{Material Net.} We use the first five convolution blocks of the ResNet-18~\cite{he2016deep}. The first layer has a filter size of $7 \times 7$ and all subsequent layer have filters of size $3 \times 3$. The number of output filters at each layer are $64, 64, 128, 256, 512$ respectively.

\noindent\textbf{Attention Net.} We use five fractionally strided convolutional layers with output filter sizes of $512, 256, 128, 64, 1$ respectively for each layer. We use filter size, stride and padding to be $4,2,1$ respectively.

%% file: supplementary/implementation_details.tex
\section{Implementation Details}
\label{sec:implementation_details}

\noindent\textbf{Input.} The input to the Visual Net and Material Net is a $128$x$128$ RGB image. We also perform image augmentation by randomly jittering color, contrast and brightness of the image.

The input to the Echo Net is a spectrogram from the simulated echoes. For obtaining the spectrogram, we first convert the time domain audio signal into Short Time Fourier Transform using Hanning window with a fixed window length, hop length and frequency points. We use a two channel audio with duration of $60$ms. 

For Replica, we use an audio signal of $44.1$kHz and convert it to a $2\times257\times166$ spectrogram using a window length of $64$, hop length of $16$ and $512$ frequency points. 

For Matterport3D, we use an audio signal of $16$kHz and convert it to a $2\times257\times121$ spectrogram using a window length of $32$, hop length of $8$ and $512$ frequency points. 

\noindent\textbf{Additional Parameters.} We train the network on both the datasets using Adam optimizer with learning rate of $1e-4$, momentum of $0.9$ and weight decay of $5e-4$. We use batch size of $128$ for Replica and $64$ for Matterport3D. 

%% file: supplementary/evaluation_metrics.tex
\section{Evaluation Metrics}
\label{sec:evaluation_metrics}

We use following metrics to evaluate our result.

We denote the predicted depth and ground truth depth as  $\hat{\mathbf{D}}(p)$ and $\mathbf{D}(p)$ for every point $p$. We further use only those points that have valid depth value, i.e. the missing values and the points having zero depth value in $\mathbf{D}$ are ignored. We denote such valid points as $|{\mathbf{D}}|$.

\begin{itemize}
    \item Root Mean Square Error:
\begin{equation}
    \sqrt{\frac{1}{|{\mathbf{D}}|}\sum_{p \in {\mathbf{D}}}|\hat{\mathbf{D}}(p) - \mathbf{D}(p)\|^2}
\end{equation} 

\item Mean absolute relative error: 
\begin{equation}
    \frac{1}{|{\mathbf{D}}|}\sum_{p \in {\mathbf{D}}}\frac{\hat{\mathbf{D}}(p) - \mathbf{D}(p)}{\hat{\mathbf{D}}(p)}
\end{equation} 

\item Mean $\log_{10}$ error: 
\begin{equation}
    \frac{1}{|{\mathbf{D}}|}\sum_{p \in {\mathbf{D}}}\log_{10} (\hat{\mathbf{D}}(p)) - \log_{10} (\mathbf{D}(p))
\end{equation} 

\item $\delta_t$ is the percentage of pixels within the error range $t$. We define the error range as mentioned below
\begin{equation}\label{eq:delta}
max(\frac{\hat{\mathbf{D}}(p)}{\mathbf{D}(p)}, \frac{\mathbf{D}(p)}{\hat{\mathbf{D}}(p)}) < t
\end{equation} 
where $t \in \{1.25, 1.25^2, 1.25^3\}$.
\end{itemize}

%% file: supplementary/qual_res.tex
\section{More Qualitative Results}
\label{sec:qual_res}
\begin{figure*}
    \vspace{-1 em}
    \centering
    \includegraphics[width=0.8\textwidth]{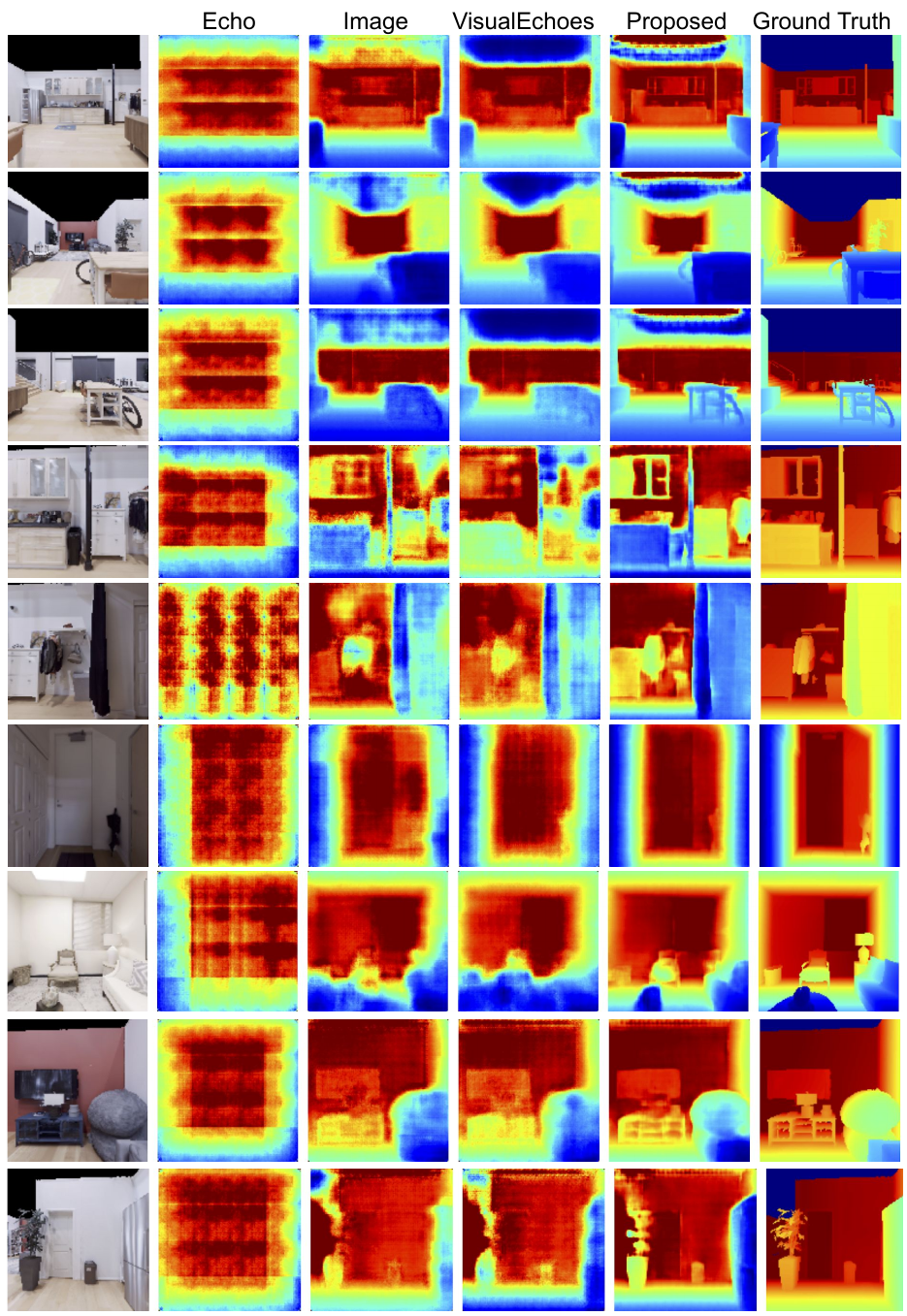}
    \caption{\textbf{Qualitative results for depth estimation on Replica dataset.} From left to right - input image, depth estimation using only echoes, depth estimation using only image, depth estimation from Visual Echoes, depth estimation using proposed method, ground truth depth map. The proposed method has better depth estimation in complicated scenes containing many objects causing frequent depth variations (\eg row $1$, row $4$). It also provides robust depth estimation along boundaries of objects (\eg rows $3$,$7$,$8$). When the individual depth estimations from image and echo are poor leading to poor depth estimation (closer to image) using Visual Echoes, while the proposed method provides closer to ground truth estimation such as cabinets (row $4$), door (row $9$) which are completely missed by other methods.}
    \label{fig:depth_pred_replica}
\end{figure*}

\begin{figure*}
    \vspace{-1 em}
    \centering
    \includegraphics[width=0.8\textwidth]{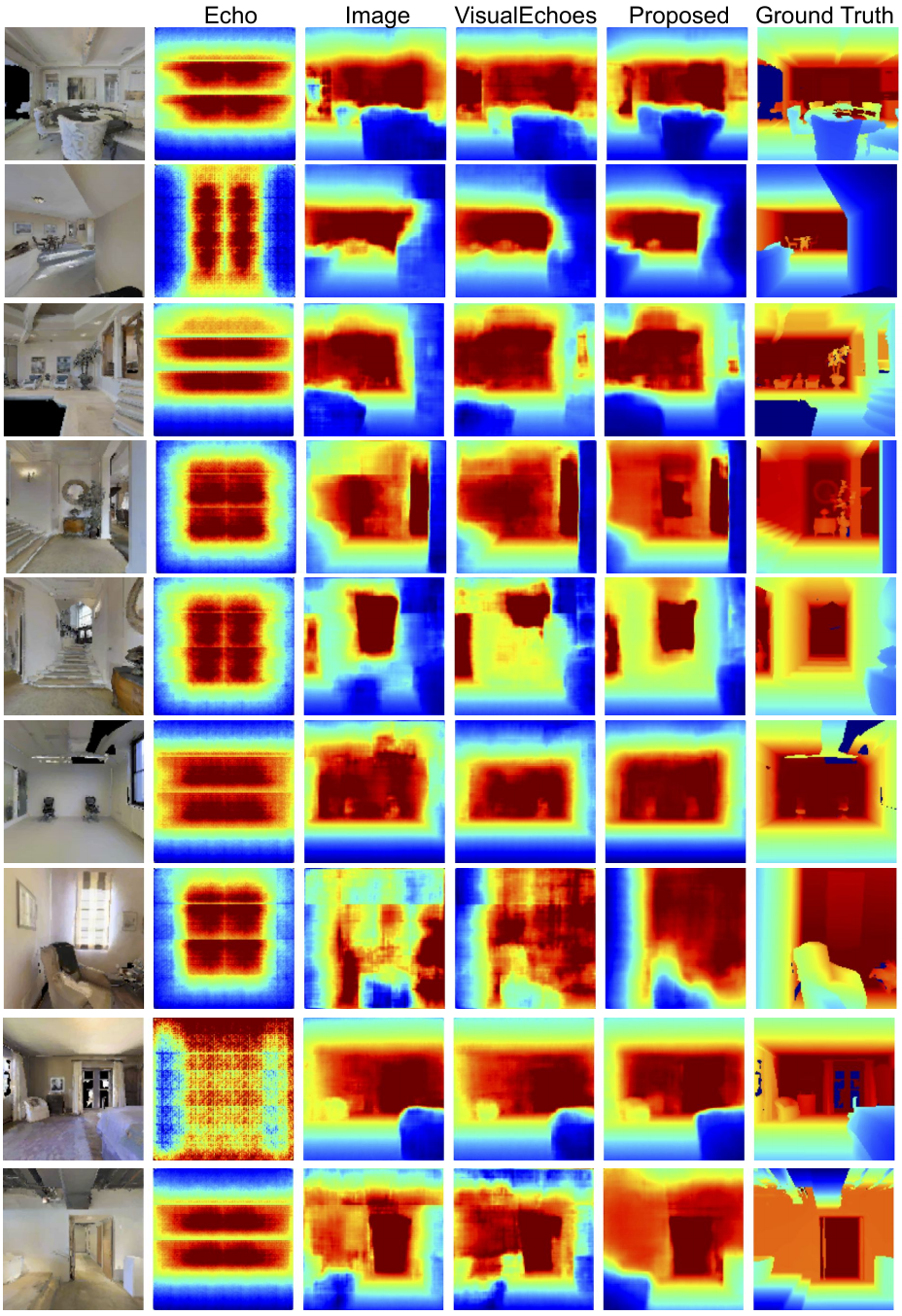}
    \caption{\textbf{Qualitative comparisons for depth estimation on Matterport3D dataset.} From left to right - input image, depth estimation using only echoes, depth estimation using only image, depth estimation from Visual Echoes, depth estimation using proposed method, ground truth depth map. We observe that the proposed method consistently provides better depth map estimation of smaller/farther objects (such as chairs \cf other methods in row $6$) and also at object boundaries (rows $1$,$4$,$5$). It also provides closer to ground truth results on illumination changes (row $7$). We also observe that when image and echo depth estimations individually yield poor results, Visual Echoes tend to perform poorly as well while the proposed method is still able to estimate better depth (row $7$).}
    \label{fig:depth_pred_mp3d}
\end{figure*}

\begin{figure}
    \centering
    \includegraphics[width=0.44\textwidth]{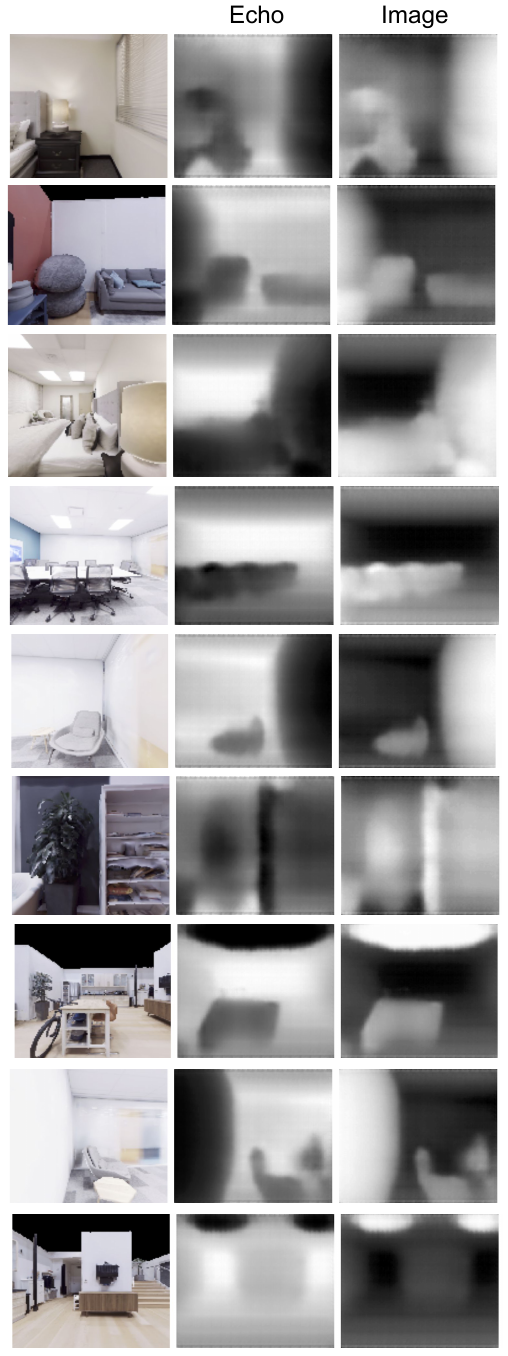}
    \caption{\textbf{Visualization of attention maps on Replica dataset.} From left to right - input image, attention map from Echo Net, attention map from Visual Net. }
    \label{fig:attention_replica}
\end{figure}

\begin{figure}
    \centering
    \includegraphics[width=0.44\textwidth]{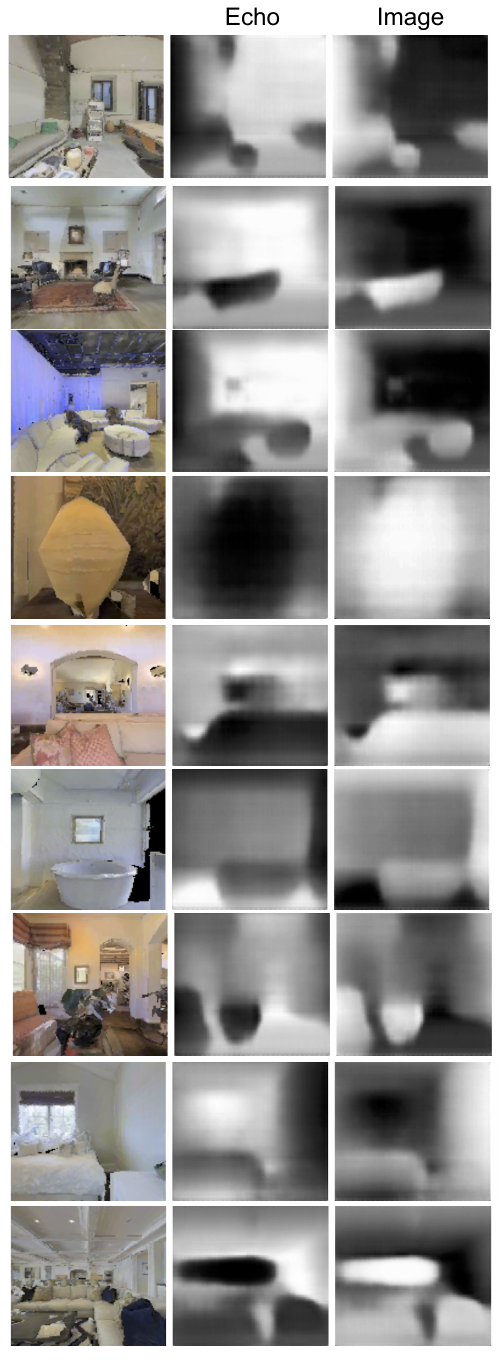}
    \caption{\textbf{Visualization of attention maps on Matterport3D dataset.} From left to right - input image, attention map from Echo Net, attention map from Visual Net.}
    \label{fig:attention_mp3d}
\end{figure}

We give more qualitative results of depth estimation using various techniques on Replica and Matterport3D datasets in \figref{fig:depth_pred_replica} and \figref{fig:depth_pred_mp3d} respectively. The visualizations of the attention maps from Echo Net and Visual Net are shown in  \figref{fig:attention_replica} (Replica) and Fig.\ref{fig:attention_mp3d} (Matterport3D).  

